\documentclass[lettersize,journal]{IEEEtran}
\usepackage{amsmath,amsfonts}
\usepackage{algorithmic}
\usepackage{algorithm}
\usepackage{array}
\usepackage[caption=false,font=normalsize,labelfont=sf,textfont=sf]{subfig}
\usepackage{textcomp}
\usepackage{stfloats}
\usepackage{url}
\usepackage{verbatim}
\usepackage{graphicx}
\usepackage{cite}
\usepackage{multirow}
\usepackage[table,xcdraw]{xcolor}
\usepackage[colorlinks,linkcolor=red]{hyperref}

\usepackage{arydshln}

\begin{document}

\title{Classification Committee for Active Deep Object Detection}

\author{Lei~Zhao, Bo~Li, Xingxing~Wei$^*$~\IEEEmembership{Member,~IEEE}
\thanks{Lei~Zhao, Bo Li, Xingxing Wei and  were at the School of Computer Science and Engineering, Beihang University, No.37, Xueyuan Road, Haidian District, Beijing,
100191, P.R. China. (E-mail: \{lzl, boli, xxwei\}@buaa.edu.cn).}
\thanks{Xingxing Wei is the corresponding author.}

}

\markboth{}%
{Shell \MakeLowercase{\textit{et al.}}: A Sample Article Using IEEEtran.cls for IEEE Journals}


\maketitle

\begin{abstract}
In object detection, the cost of labeling is much high because it needs not only to confirm the categories of multiple objects in an image but also to accurately determine the bounding boxes of each object. Thus, integrating active learning into object detection will raise pretty positive significance. In this paper, we propose a classification committee for active deep object detection method by introducing a discrepancy mechanism of multiple classifiers for samples' selection when training object detectors. The model contains a main detector and a classification committee. The main detector denotes the target object detector trained from a labeled pool composed of the selected informative images. The role of the classification committee is to select the most informative images according to their uncertainty values from the view of classification, which is expected to focus more on the discrepancy and representative of instances. Specifically, they compute the uncertainty for a specified instance within the image by measuring its discrepancy output by the committee pre-trained via the proposed Maximum Classifiers Discrepancy Group Loss (MCDGL). The most informative images are finally determined by selecting the ones with many high-uncertainty instances. Besides, to mitigate the impact of interference instances, we design a Focus on Positive Instances Loss (FPIL) to make the committee the ability to automatically focus on the representative instances as well as precisely encode their discrepancies for the same instance. Experiments are conducted on Pascal VOC and COCO datasets versus some popular object detectors. And results show that our method outperforms the state-of-the-art active learning methods, which verifies the effectiveness of the proposed method.
\end{abstract}


\section{Introduction}
\IEEEPARstart{D}{eep} learning technology is developing very fast in the field of object detection \cite{ssd,fasterrcnn,redmon2017yolo9000,wang2020ship,yuan2022translation,bo2021ship}, it is necessary not only to design appropriate models and training strategies but also to acquire a great amount of training data. Generally, the cost of obtaining many labeled samples is very expensive. For this problem, active learning comes into being, the aim of active learning is to select the most informative samples for a specific task. In computer vision, previous research for active learning focus on image classification \cite{wu2020multi,li2013adaptive,gilad2005query,cao2020hyperspectral,sinha2019variational,yan2016image,wang2016cost}. The labeling work for image classification is relatively easy compared to labeling for object detection because an image for object detection not only contains the classification labels of multiple objects but also needs to precisely label the bounding boxes of all objects in the image. Thus, active learning techniques are increasingly needed in the field of object detection \cite{choi2019gaussian,devries2018learning, Yuan_2021_CVPR,qiu2020hierarchical,haussmann2020scalable,li2021deep}.
\begin{figure*}[htp]
\centering
\includegraphics[scale=0.75]{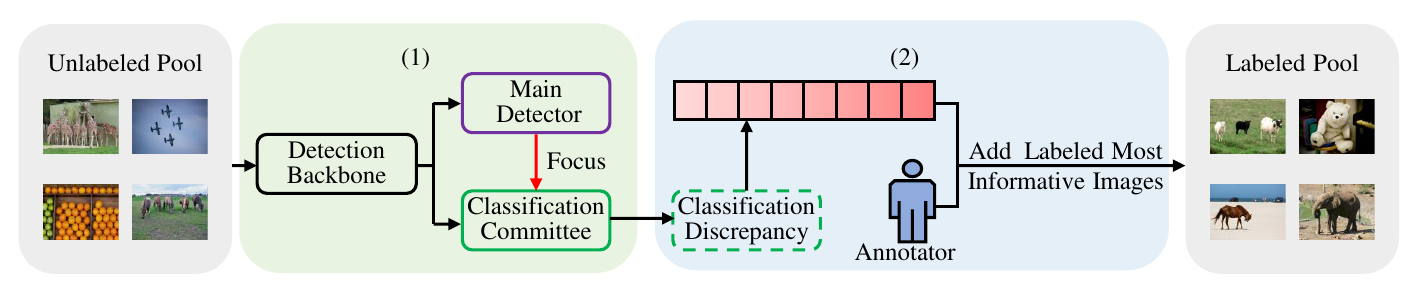} 
\caption{The flow diagram of the proposed active learning method for object detection. (1) The object detection model with a main detector and a classification committee is well-trained, using MCDGL and FPIL to maximize positive instances of uncertainty in the training phase of each cycle. (2) The classification discrepancy of the classification committee is calculated as classification uncertainty. These uncertainties of instances in an image are aggregated to the uncertainty of the image. The most informative sample is selected for annotation and added to the labeled pool for the next cycle.}
\label{framework}
\end{figure*}

The key idea of active learning is how to evaluate the most informative samples from unlabeled data for the target model \cite{holub2008entropy,beluch2018power,yuan2019multi,liu2022survey,liao2016visualization,wang2023active}. Current active learning approaches mainly aim at the classification task and utilize an uncertainty-based strategy to measure the output of the target model \cite{liao2016visualization,siddiqui2020viewal,wang2018uncertainty,lughofer2017online}. 
For anchor-based object detection, because one head of the object detector is to determine the classification label, some traditional strategies for image classification can be introduced  on the classification cue of the detector \cite{haussmann2020scalable,roy2018deep}. The reasons that classification-based active learning methods can work well for object detectors are: (1) The classifier has the higher spatial sensitivity than the localization cue and can better distinguish the complete object and part of an object through observing the output feature maps of both localization and classification \cite{wu2020rethinking}. (2) The localization loss only works when an anchor box is an object \cite{ssd,redmon2017yolo9000,girshick2015fast} when lots of inaccurate boxes are obtained in the prediction stage, they may interfere with the uncertainty evaluation. As a comparison, the classification cue is less affected than the localization cue.

Although effective, the current classification-based active learning methods usually use one classifier to evaluate and sum the uncertainty of predicted multiple objects as an image score \cite{roy2018deep,ranganathan2017deep}. There are four disadvantages to such a process:
(1) one classification's output is not considered to evaluate samples' uncertainty from multiple perspectives in the feature space, because one decision boundary trained by the labeled samples is not really accurate for the unlabeled samples and misses the potential information. (2) The distribution difference between labeled and unlabeled data sets is not considered. When directly extending the model trained on the labeled data set to the uncertainty evaluation of the unlabeled data set, it will result in inaccurate predictions in the unlabeled data set. (3) The detector is a deep model but the current selector utilizes the handcrafted feature or rule-based metric. It is a static system with poor portability and knowledge bottleneck, which does not use learning-based assessment from learned features of itself, resulting in the loss of information evaluation. (4) Aggregating uncertain instances directly cannot represent the uncertainty of the image effectively, because it contains lots of background information which is extremely imbalanced for positive instances \cite{Yuan_2021_CVPR} and this phenomenon can destroy the consistency of foreground uncertainty and image uncertainty, hindering the selection of informative samples.

\begin{figure*}[htp]
\centering
\includegraphics[scale=0.75]{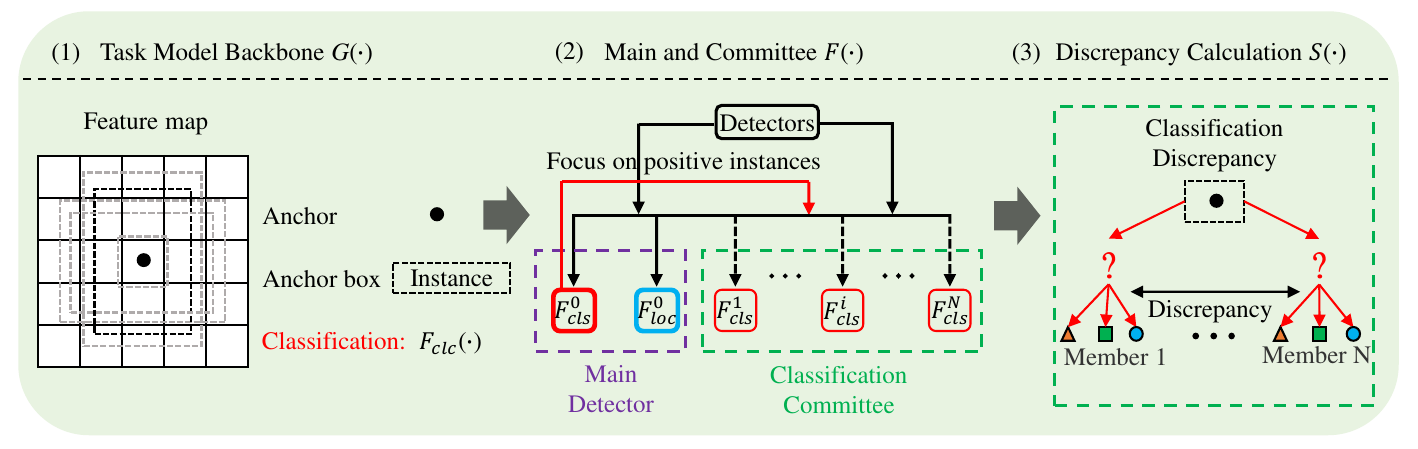} 
\caption{The details of the active learning object detection. (1) An anchor-based object detection, the features corresponding to anchor boxes are defined as an instance. (2) The committee architecture contains a main detector and a classification committee. The classifier of the main detector can provide background instances classification scores to focus on more uncertain positive instances. (3) The discrepancy of the classification committee is classification uncertainty.} 
\label{detector}
\end{figure*}

In order to solve the above problems, we propose a classification committee for active deep object detection method, selecting informative images from the unlabeled pool by considering the discrepancy of multiple classifiers' decision boundaries. 
The model architecture is shown in Figure \ref{framework}, a classification committee composed of multiple classifiers evaluates the uncertainty of the image from multiple perspectives for sample selection to improve the performance of the main detector faster. Maximum discrepancy learning in the committee is the deep feature, to expect maximum discrepancies between pairwise members in the committee, the discrepancy of learning needs to traverse all possible member combinations, which is inefficient. So, we propose Maximum Classifiers Discrepancy Group Loss (MCDGL) to train the committee on the unlabeled pool, which maximizes the discrepancy by keeping each member away from their center at the same time. 
Meanwhile, for the committee to focus as much as possible on the uncertainty of more representative positive instances, we propose the Focus on Positive Instances Loss (FPIL) that utilizes the scores of the main classifier weighted to MCDGL, as shown in Figure \ref{detector}, narrowing the discrepancies of interference instances and filtering out background information in instance selection for image uncertainty estimation. By highlighting the uncertainty of the representative instances as the uncertainty of the image, the most informative samples are added to the training set to train the detector in the next cycle. 

The contributions of this paper include:
\begin{itemize}
\item We propose a classification committee for active deep object detection method, which includes a main detector and a classification committee.  
The classification committee is used for the uncertainty estimation of instances, and the uncertainty sum of representative instances is used as the information content of the image for image selection.   

\item To effectively learn the discrepancy of the classification committee, we propose Maximum Classifiers Discrepancy Group Loss (MCDGL). To suppress the information of the interference instances, we propose Focus on Positive Instances Loss (FPIL), which can highlight the uncertainty of representative instances.

\item We have verified the effectiveness of MCDGL and FPIL for selecting the most informative images from the unlabeled pool, it is a great improvement over the state-of-the-art methods. Especially for PASCAL VOC in some cycles, our method can enhance the mAP by nearly 6.0\% and nearly 2.0\% for using RetinaNet and SSD, respectively.
\end{itemize}

The remainder of this paper follows a structured approach. In Section \ref{related}, we conduct a comprehensive review of existing literature concerning active learning for deep learning, as well as its applications in the context of object detection. Moving on to Section \ref{methods}, we provide a detailed explanation of the proposed method, outlining its key components and mechanisms. Subsequently, in Section \ref{exp}, we present the experimental results obtained from applying the proposed method and conduct a thorough analysis of these results. Finally, in Section \ref{con}, we conclude the paper by summarizing the main findings and discussing the potential implications and future directions of this research.

\section{Related Work}
\label{related}
\subsection{Active learning for deep learning}
Deep learning generally requires large amounts of data, which drives the development of active learning. Common active learning strategies can be used directly in deep learning, which can calculate a posterior probability distribution whose shape can reflect the uncertainty of the sample. For example, least confidence selects minimum confidence of maximum classification score from unlabeled pooling \cite{agrawal2021active,zhu2010confidence}. Also, relying on entropy measures information of unlabeled data \cite{siddiqui2020viewal,allotey2021entropy}. Meanwhile, some researchers also focus on data distribution, Core-Set \cite{sener2017active,kim2022defense} estimates the diversity of a set, utilizing the model trained by labeled samples to obtain the diversity of the unlabeled samples. The clusters center \cite{COLETTA2019150,wang2017active} for the probability distribution of unlabeled samples is the most representative sample. The sample with the largest change in expected parameters and predicted losses is usually the sample with the most effective information \cite{samat2019edge,venkatesh2020ask,kading2016active}. In addition, an adversarial network is trained to discriminate whether a latent space of variational autoencoder is from a label sample pooling or unlabeled pooling \cite{sinha2019variational}. Learning loss \cite{yoo2019learning} adds additional modules from the feature layer to learn uncertainty which is predicted loss, it is the deep feature for uncertainty estimation, but only the model trained by labeled samples is used to analyze the uncertainty of unlabeled samples. An image for object detection contains not only different categories and quantities of objects but also a large amount of background information. Neither collecting uncertain instances nor using the information of the intermediate feature layer to predict the information amount can directly represent the information amount of the image and they cannot be used directly for active deep object detection effectively.

\subsection{Active Learning for Object Detection}
In recent years, in addition to simply porting the method of the classification task, some researchers have introduced novel active learning into object detection in the field of deep learning \cite{brust2018active,kao2018localization,haussmann2020scalable,choi2019gaussian,haussmann2020scalable}. Actually, learning loss is also applied to object detection \cite{yoo2019learning}, but the global average pooling can not represent the sum of uncertainties of multiple objects of the entire image. Also, pixel-level uncertainty can aggregate image-level uncertainty, according to the image-level uncertainty to select the most informative image \cite{aghdam2019active}. Query by committee consists of multiple decision makers and can search for the samples that have the most disagreement of different perspectives \cite{seung1992query}. Querying images also utilize the disagreement between the convolution layers in the SSD \cite{roy2018deep}, but the method also designs more complex handcraft feature or rule-based metric rather than automatically learned deep features. These methods also ignore the uncertainty caused by a large amount of background interference information, and the information of the middle layer or various aggregation methods are not fully representative of the information of the image and also are only trained with labeled samples for information evaluation of unlabeled samples. MI-AOD \cite{Yuan_2021_CVPR} combines semi-supervised and pseudo-labeling techniques to make it impossible to determine whether active learning plays a role in the end because both of these techniques are involved in backbone promotion. Two classifiers without the main detector interfere with each other, and multi-instance learning focuses more on whether there is an object, and the score of whether an instance is background is not accurate enough. Moreover, the training process of MI-AOD is more complex and needs repeated back-and-forth adversarial training. 

In contrast, our approach utilizes committee discrepancy to evaluate the value of images, which can observe the uncertainty of the sample from multiple decision boundaries. The committee needs to be trained with unlabeled samples, it is independent of the main detector so as not to affect the performance of the main detector, and the training process is also independent rather than adversarial training. Uncertainty learning is also a deep feature suitable for the uncertainty assessment of deep models.

\section{Methods}
\label{methods}
\subsection{Overview}
In the section, for active learning of object detection, a large unlabeled set is defined as $\mathcal{X}_U^{Q}$. The superscript $Q$ represents the number of unlabeled images. 
$K_p$ is the number of samples in the labeled pool after the $p$-th cycle of active learning, and some of the most informative samples from the unlabeled pool are added to the labeled pool at the next cycle. Thus, for the $p$-th cycle, the labeled pool is $\mathcal{X}_L^{K_p}$ and the unlabeled pool is $\mathcal{X}_U^{Q-K_p}$. The active learning cycle stops when the budget runs out or reaches a certain precision. For each cycle, the model needs to be retrained, and $\mathcal{W}_p$ is the $p$-th cycle trained model.

In the study, we focus on anchor-based object detection for active learning. These methods present a large number of anchor boxes, and an anchor box is defined as an instance that contains anchor features \cite{lin2017focal}. To evaluate the uncertainty of unlabeled instances in an unlabeled image by using the trained object detection model, we propose a classification committee as a set of auxiliary classifiers on the original detector, which does not affect the performance of the main detector by detaching the gradient between the feature extractor and the committee, as shown in Figure \ref{framework}. The structure is different from MI-AOD, our approach does not involve and apply complex adversarial training. The discrepancy of the classification committee represents the uncertainty of instances in the image, and we design MCDGL to learn the discrepancy of these classifiers that maximize instances of discrepancy in the committee. For an image, lots of instances belonging to the background and interference information can affect the uncertainty estimation of the whole image, which can cause these most uncertain instances not to represent the true amount of information in the image. Thus, we propose FPIL that can focus more on the discrepancy of positive instances, which utilizes the main classification score to weight MCDGL and requires no additional image-level calculation, as shown in Figure \ref{focus}.

\subsection{Classification Committee for Uncertainty}
\begin{figure}[htp]
\centering
\includegraphics[scale=0.45]{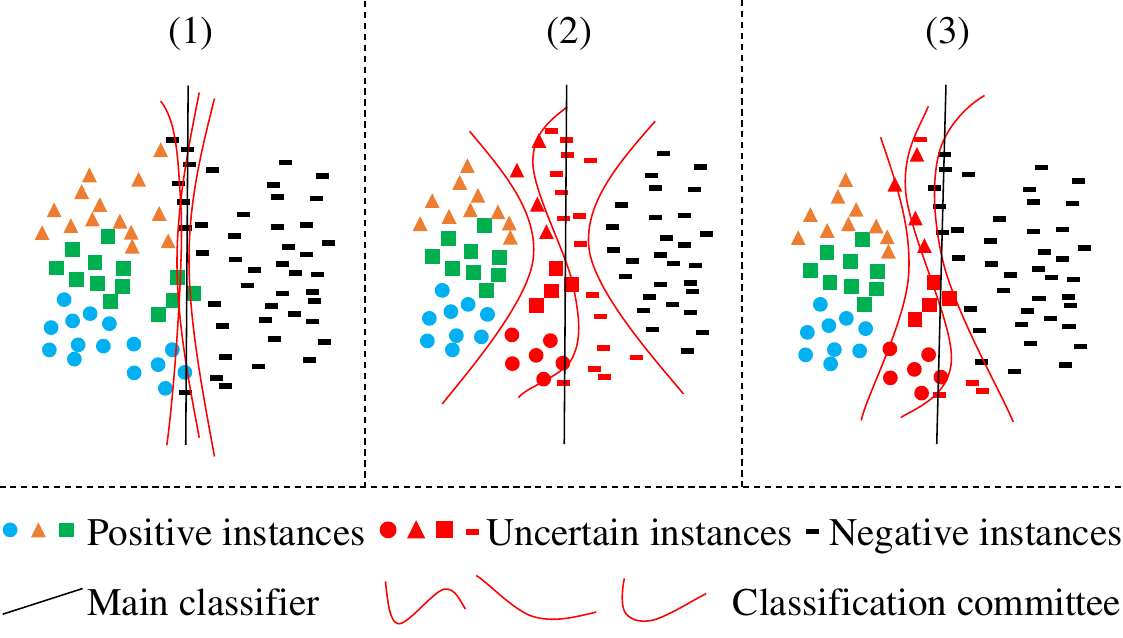} 
\caption{The schematic diagram of instance distribution. (1) Only train the original detection loss for the main classifier and classification committee. (2) Maximizing classification discrepancy, the red uncertain instances contain lots of positive and background instances. (3) FPIL can pay more attention to the discrepancy of positive instances.} 
\label{focus}
\end{figure}

For the anchor-based object detection framework, the backbone network of feature extraction is defined as $\mathit{G}(\cdot)$ and its parameters are $w_G$. Inquired by query-by-committee for active learning\cite{gilad2005query}, we hope that committee decisions and discrepancies between members of the committee can be used as a basis to evaluate the amount of information. Thus, we retain the original detector as the main detector, and add a classification committee, as shown in Figure \ref{framework}. The detectors are defined as $\mathit{F}(\cdot)$, and their parameters are $w_F$. The classification and the localization of the main detector are defined as $\mathit{F}_{cls}^0$ and $\mathit{F}_{loc}^0$, respectively. Analogously, we define the classification committee set as \{$\mathit{F}_{cls}^i,i=1,\cdots,N$\}. $N$ represents the number of members in the classification committee. $w_F=\{w_F^0,w_F^{com}\}$, $w_F^0$ is the main detector parameters and $w_F^{com}$ is the classification committee parameters. The classification committee trained by the unlabeled pool mainly wants their predictions to be more discrepancy for an uncertain instance, which does not affect the detection performance of committees trained with the labeled pool. Thus, the main detector trained by the labeled pool is expected to be more accurate. In object detection, the total number of instances corresponding to different feature layers for an image $\mathit{X}$ is very large, and the instances can be represented as \{$\mathit{x}_j,j=1,\cdots, T$\} = $G(X)$, where $T$ is the total number of instances for the image. The actual labels of these instances are defined as \{ $\mathit{y}_j^{cls},j=1,\cdots,T$ \}. For the predicted values of main detector,$\mathit{y}_j^{\mathit{f}_{cls}^0}=\mathit{F}_{cls}^0(\mathit{x}_j)$ and $\mathit{y}_j^{\mathit{f}_{loc}^0}=\mathit{F}_{loc}^0(\mathit{x}_j)$. Similarly, for the classification committee, $\mathit{y}_j^{\mathit{f}_{cls}^i}=\mathit{F}_{cls}^i(\mathit{x}_j)$. For the image $\mathit{X}$ of batch size $\mathcal{B}$ samples in the labeled pool,
$\mathcal{X}_L^{\mathcal{B}}\in \mathcal{X}_L^{K_p}$, the main detector can be optimized by the following loss function, as 
\begin{equation}
\begin{split}
\mathop{\arg\min}\limits_{w_G,w_F^0} \mathit{L}_{main}(\mathit{X})=&\frac{1}{T}\sum_{j=1}^T(FL(y_j^{\mathit{f}^0_{cls}},y_j^{cls})\\
&+SmoothL1(y_j^{\mathit{f}^0_{loc}},y_j^{loc})),
\label{gs1}
\end{split}
\end{equation}
where $FL(\cdot)$ is the focal loss for the classification of object detection and $SmoothL1(\cdot)$ is used to optimize the bounding box regression\cite{lin2017focal}. For the labeled pool in the $p$-th cycle, the loss function for the main detector is defined as
\begin{equation}
\mathop{\arg\min}\limits_{w_G,w_F^0}
\mathcal{L}_{main}^p=\frac{1}{\mathcal{B}}\sum_{\mathit{X} \in \mathcal{X}_L^{\mathcal{B}}}\mathit{L}_{main}(\mathit{X}).
\label{gs2}
\end{equation}
For the classification committee on the labeled set, the committee also needs to be properly trained, thus the loss function for the classification committee for image $\mathit{X}$ is defined as
\begin{equation}
\mathop{\arg\min}\limits_{w_F^{com}}
\mathit{L}_{com}(\mathit{X})=\frac{1}{N}\frac{1}{T}\sum_{i=1}^N\sum_{j=1}^TFL(y_j^{\mathit{f}^i_{cls}},y_j^{cls}),
\label{gs3}
\end{equation}
where $N$ is the number of members. For the labeled pool in the $p$-th cycle, the loss function for committees is defined as
\begin{equation}
\mathop{\arg\min}\limits_{w_F^{com}}
\mathcal{L}_{com}^p=\frac{1}{\mathcal{B}}\sum_{X \in \mathcal{X}_L^{\mathcal{B}}}\mathit{L}_{com}(\mathit{X}).
\label{gs4}
\end{equation}
Both the main detector and classification committee can perform well on the labeled pool, and their output features are similar, as shown in Figure \ref{focus} (1). The main detector is mainly used for detection work and the classification committee is mainly used to calculate the discrepancy of the unlabeled instances. The main classifier is not trained with unlabeled samples, so the discrepancy calculation process will not be added, and the classification committee will not affect the performance of the main detection. Therefore, the committee needs to maximize uncertainty for instances on the unlabeled pool, as shown in Figure \ref{detector} and Figure \ref{focus} (2). Thus, the discrepancy among the members is as follows
\begin{equation}
\mathit{D}_{ins}(x_j)=\frac{1}{N(N-1)}\sum^N_{i=1}\sum^N_{t=1,t\neq i}d_{cls}(y_j^{\mathit{f}^i_{cls}},y_j^{\mathit{f}^t_{cls}}),
\label{gs5}
\end{equation}
where $d_{cls}(y_j^{\mathit{f}^i_{cls}},y_j^{\mathit{f}^t_{cls}})$ is the discrepancy between the $i$-th member and the $t$-th member of classification committee.  $d_{cls}(\cdot)$ can adopt L1 distance or L2 distance, and we select L2 instance in this paper. But Eq.(\ref{gs5}) is complex and expensive to implement because it has to traverse all member pairs in the committee, resulting time complexity of $N^2$. To solve this problem, inspired by \cite{liu2019group}, using L2 distance as $d_{cls}(\cdot)$, because $||y_j^{\mathit{f}^i_{cls}}-y_j^{\mathit{f}^t_{cls}}||=0$ when $i=t$, the modified formula of $\mathit{D}_{ins}(x_j)$ is as follows 
\begin{equation}\small
\begin{split}
&\mathit{D}_{ins}(x_j)\\
&=\frac{1}{N^2}\sum_{i=1}^N\sum_{t=1}^N||y_j^{\mathit{f}^i_{cls}}-y_j^{\mathit{f}^t_{cls}}||^2\\
&=\frac{1}{N^2}\sum_{i=1}^N\sum_{t=1}^N(||y_j^{\mathit{f}^i_{cls}}||^2-2y_j^{\mathit{f}^i_{cls}}\cdot y_j^{\mathit{f}^t_{cls}}+||y_j^{\mathit{f}^t_{cls}}||^2)\\
&=\frac{1}{N^2}(\sum_{i=1}^NN||y_j^{\mathit{f}^i_{cls}}||^2-\sum_{i=1}^N2y_j^{\mathit{f}^i_{cls}}\cdot\sum_{t=1}^Ny_j^{\mathit{f}^t_{cls}}+\sum_{t=1}^NN||y_j^{\mathit{f}^t_{cls}}||^2)\\
&=\frac{2}{N^2}(\sum_{i=1}^NN||y_j^{\mathit{f}^i_{cls}}||^2-\sum_{i=1}^Ny_j^{\mathit{f}^i_{cls}}\cdot N\overline{y}_j^{\mathit{f}_{cls}})\\
&=\frac{2}{N}(\sum_{i=1}^N||y_j^{\mathit{f}^i_{cls}}||^2-\sum_{i=1}^N2y_j^{\mathit{f}^i_{cls}}\cdot \overline{y}_j^{\mathit{f}_{cls}}+\sum_{i=1}^N||\overline{y}_j^{\mathit{f}_{cls}}||^2)\\
&=\frac{2}{N}\sum_{i=1}^N||y_j^{\mathit{f}^i_{cls}}-\overline{y}_j^{\mathit{f}_{cls}}||^2.
\end{split}
\label{gs6}
\end{equation}
In Eq.(\ref{gs6}), $y_j^{\mathit{f}^i_{cls}}\in \mathbb{R}^{1\times C}$ where $C$ is the number of classes, $\overline{y}_j^{\mathit{f}_{cls}}$ is the mean output of classification committee. This paper \cite{liu2019group} needs to narrow the distance among pairs of samples from different feature groups. Inspired by \cite{seung1992query}, maximizing the discrepancy of the committee, the information gain for the sample query can be improved. Thus, we also hope that maximization Eq.(\ref{gs6}) is to learn the sum of the discrepancy of all of these pairs of members, thus the formula is named Maximum Classifiers Discrepancy Group Loss (MCDGL). 
For the image $\mathit{X}$ of batch size $\mathcal{B}$ samples in the unlabeled pool, the maximizing discrepancy of the image is as follows
\begin{equation}
\mathop{\arg\max}\limits_{w_F^{com}}
\mathit{D}_{img}(X)=\frac{1}{T}\sum_{j=1}^T\mathit{D}_{ins}(x_j).
\label{gs7}
\end{equation}
Meanwhile, the unlabeled pool needs to maximize discrepancy, $\mathcal{X}_U^{\mathcal{B}}\in\mathcal{X}_U^{Q-K_p}$. For the $p$-th cycle, the loss formula is as follows
\begin{equation}
\mathop{\arg\max}\limits_{w_F^{com}}
\mathcal{D}_{com}^p=\frac{1}{\mathcal{B}}\sum_{X \in \mathcal{X}_U^{{\mathcal{B}}} }\mathit{D}_{img}(X).
\label{gs8}
\end{equation}
Finally, the total loss function of the $p$-th cycle for the classification committee active learning framework of object detection is as follows
\begin{equation}
\mathop{\arg\min}\limits_{w_G,w_F}
\mathcal{T}_p=\mathcal{L}_{main}^p+\mathcal{L}_{com}^p
-\lambda\mathcal{D}_{com}^p,
\label{gs9}
\end{equation}
where $\lambda $ is regularization hyper-parameters. 

\subsection{Enhancement of Uncertain Positive Instances}
In the last section, when training is complete for the previous cycle, for an unlabeled image, the top-$Z$ uncertain instance is selected according to the value order of Eq.(\ref{gs6}) as the uncertainty for the image. But the number of instances containing background information is much larger than the number of instances containing object information. The top-$Z$ selected uncertain instances can not really represent the amount of information of an image, and the selected instances are expected to contain more positive instances of uncertainty. The two problems we need to solve are the suppression of background uncertainty and the reduction of the discrepancy of very certain positive instances. The classifier of the main detector can predict the probability score of an instance or the probability score of the background class, and we define the probability score of the background instance as $w^b_j\in\mathit{y}_j^{\mathit{f}_{cls}^0}$. We design the Focus on Positive Instances Loss (FPIL) that Eq.(\ref{gs7}) is modified as follows
\begin{equation}
\widetilde{\mathit{D}}_{img}(X)=\frac{1}{T}\sum^T_{j=1}(1-w^b_j)^{\gamma}\mathit{D}_{ins}(x_j),
\label{gs10} 
\end{equation}
where $\gamma$ is an adjustment factor, $\gamma>0$. By optimizing Eq.(\ref{gs10}) for an image, the greater the probability of instances classification of background, the smaller the discrepancy between members. Meanwhile, for very sure positive instances, $\mathit{D}_{ins}(x_j)$ is small,  $(1-w^b_j)^{\gamma}\mathit{D}_{ins}(x_j)$ is also small. In other words, the smaller the discrepancy of committees for an instance, the more certain it is. Only instances with smaller background category scores and more uncertain instances are positive instances of uncertainty required for active learning. As shown in Figure \ref{focus}, the diagram shows how different classifiers change in the feature space. $\widetilde{\mathit{D}}_{img}(X)$ is used by Eqs.(\ref{gs7}), (\ref{gs8}), (\ref{gs9}) to obtain $\widetilde{\mathcal{D}}^{p}_{com}$. The final loss function is as follows
\begin{equation}
\mathop{\arg\min}\limits_{w_G,w_F}
\widetilde{\mathcal{T}}_p =\mathcal{L}^p_{main}+\mathcal{L}^p_{com}-\lambda\widetilde{\mathcal{D}}^p_{com}.
\label{gs11}
\end{equation}
Finally, during the entire training phase of the model, in order not to let the gradient update of the classification committee affect the feature extraction and the features of the main detector, the gradient returned by the classification committee is detached from the backbone. For experimental fairness, focus only on active learning itself. The training process is not complex adversarial training and is not suitable for semi-supervised learning.
\begin{algorithm}[tb]
\caption{Active Learning for Object Detection}
\label{alg:algorithm}
\textbf{Input}: The number of members in committee: $N$; The number of selected uncertain instances: $Z$; The number of the active cycle: $P$; Initially unlabeled images: $\mathcal{X}_U^Q$.\\
\textbf{Initialization}: Parameters $w_G$ and $w_F$ in $G(\cdot)$ and $F(\cdot)$; Initial model $\mathcal{M}_0$; Initial labeled and unlabeled pool are $\mathcal{X}_L^{K_0}$ and $\mathcal{X}_U^{Q-K_0}$.
\begin{algorithmic}[1] 
\STATE Let $p=0$.
\WHILE{$p \leq P$}
\IF {Using $\mathcal{X}_L^{K_p}$}
\STATE Update $w_G$ and $w_F$ by Eqs.(\ref{gs2}) and (\ref{gs4})
\ENDIF
\IF {Using $\mathcal{X}_U^{Q-K_p}$}
\STATE Update $w_F^{com}$ by Eq.(\ref{gs10})

\ENDIF
\STATE Evaluate the average accuracy of the model, $mAP$\\
\STATE Calculate committee discrepancy by Eq.(\ref{gs6})
\STATE Calculate $S(X)$ of image by Eqs.(\ref{gs12}) and (\ref{gs13})
\STATE Label the $K_{p+1}-K_{p}$ images that are most worthy
\STATE Update $\mathcal{X}_L^{K_p}$ to $\mathcal{X}_L^{K_{p+1}}$\\
\STATE Update $\mathcal{X}_U^{Q-K_p}$ to $\mathcal{X}_U^{Q-K_{p+1}}$\\
\STATE Obtain $\mathcal{W}_p$
\STATE $p=p+1$. 
\ENDWHILE
\STATE \textbf{return} $\mathcal{W}_P$
\end{algorithmic}
\label{al}
\end{algorithm}

\begin{figure*}[htbp]
\centering
    \begin{minipage}[t]{0.33\linewidth}
        \centerline{\includegraphics[scale=0.35]{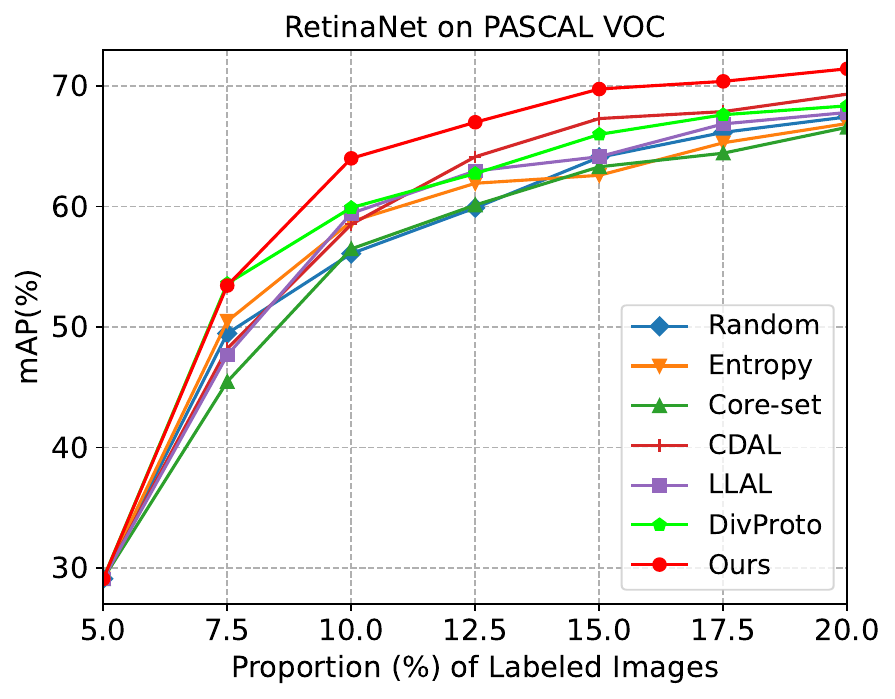}}
         \centerline{\quad(\small 1)}
    \end{minipage}%
    \begin{minipage}[t]{0.33\linewidth}
        \centerline{ \includegraphics[scale=0.35]{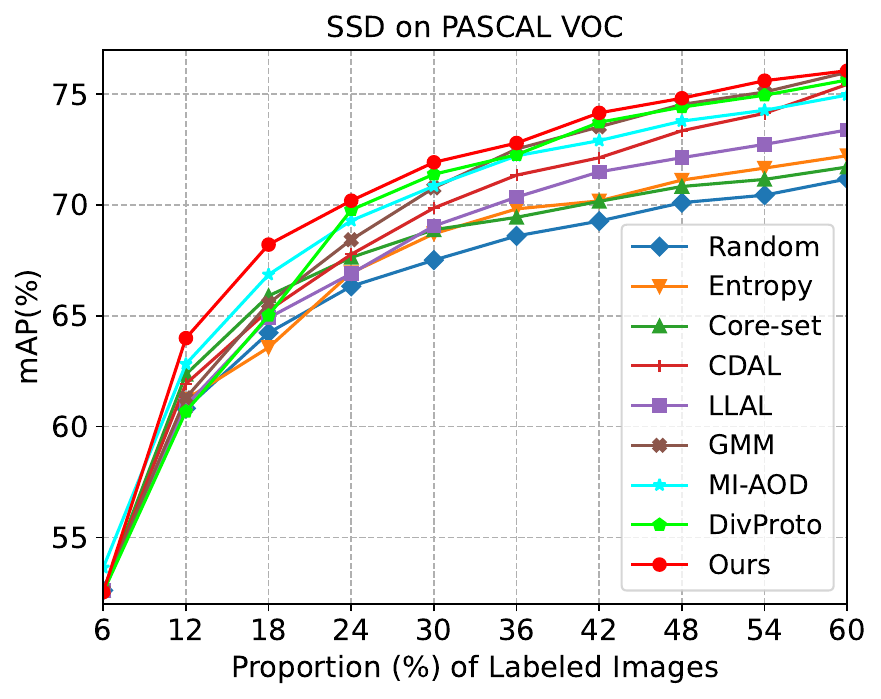}}
        \centerline{\quad(\small 2)}
    \end{minipage}
    \begin{minipage}[t]{0.33\linewidth}
        \centerline{\includegraphics[scale=0.35]{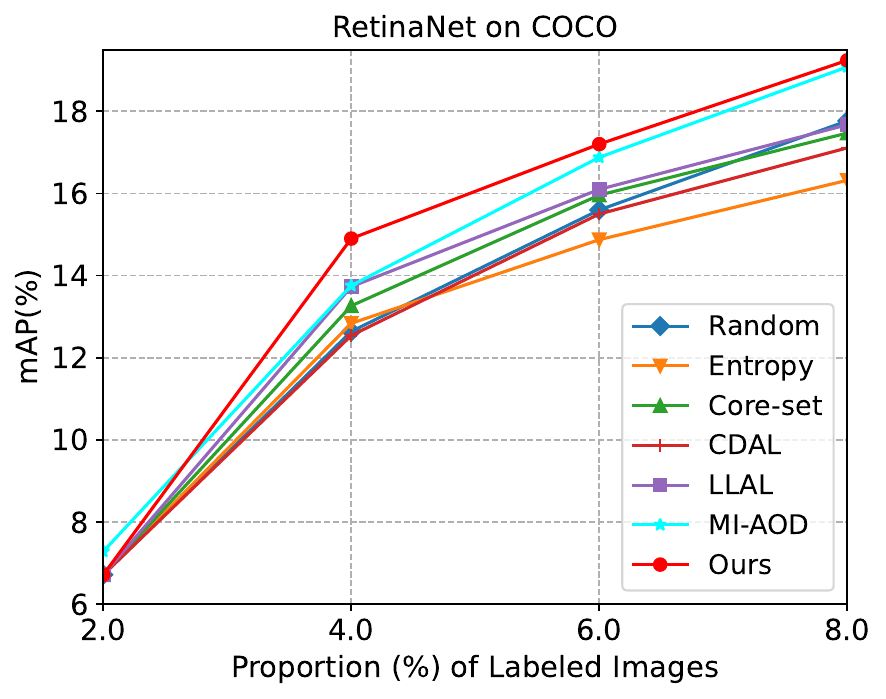}}
        \centerline{\quad(\small 3)}
    \end{minipage}    
    \caption{ Performance comparison of active object detection methods. (1) On PASCAL VOC using RetinaNet. (2) On PASCAL VOC using SSD. (3) On MS COCO using RetinaNet.}
    \label{fig1}
\end{figure*}

\subsection{Images Selection for Active Learning}
By optimizing Eq.(\ref{gs11}) for the whole model, we can select the top-$Z$ uncertain instances in an image according to Eq.(\ref{gs6}), the sum of these uncertainties represents the uncertainty of the image. The uncertainty information score $\mathit{S}(X)$ is as follows:
\begin{equation}
\mathit{S}(X)=\frac{1}{Z}\sum^Z_{i=1}s_i,
\label{gs12}
\end{equation}
where $s_i$ is the score of the $i$-th uncertain instance in the sequence obtained after the uncertainty of all instances is calculated by Eq.(\ref{gs6}) and sorted in descending order. The formula is as follows:
\begin{equation}
\begin{split}
S_T&=\{s_i,i=1,\cdots,T\}\\
&=DescendSort(D_{ins}(x_j),j=1,\cdots,T).
\end{split}
\label{gs13}
\end{equation}
In Eq.(\ref{gs13}), $S_T$ is a sequence set that has $T$ elements. According to $\mathit{S(X)}$, we can choose which is more worthy of labeling from the unlabeled pool. The process of the proposed method is shown in Algorithm \ref{al}.

\renewcommand{\arraystretch}{1.8}
\begin{table*}[t] 
\centering
\begin{center}
\label{tab1}
\caption{Module ablation on PASCAL VOC using RetinaNet and SSD for MCDGL and FPIL.}
\begin{tabular}{p{1.8cm}<{\centering}|p{1.4cm}<{\centering}|p{1.4cm}<{\centering}| p{1.2cm}<{\centering} p{1.2cm}<{\centering} p{1.2cm}<{\centering} p{1.2cm}<{\centering} p{1.2cm}<{\centering} p{1.2cm}<{\centering} p{1.2cm}<{\centering} }
\hline
\multirow{4}*{RetinaNet} &\multicolumn{2}{c|}{Training} &\multicolumn{7}{c}{mAP (\%) on Proportion (\%) of Labeled Images.}  \\
\cline{2-10}
~ & MCDGL & FPIL &5.0 &7.5 &10.0 &12.5 &15.0 & 17.5 & 20.0 \\
\cline{2-10}
 ~ & \checkmark &~ &29.09 &51.51 &62.01 &65.45 &67.07  &70.03 &70.87\\
 ~ & \checkmark  &\checkmark &29.09 &\textbf{53.43} &\textbf{63.98} &\textbf{66.96} &\textbf{69.73} &\textbf{70.37} &\textbf{71.41}\\
 \hline
\multirow{4}*{SSD} &\multicolumn{2}{c|}{Training} &\multicolumn{7}{c}{mAP (\%) on Number of Labeled Images.}  \\
\cline{2-10}
~ &MCDGL &FPIL &6 &12 &18 &24 &30 &36 &42\\
\cline{2-10}
 ~ & \checkmark & ~   &52.62 &63.48 &67.61 &70.03 &71.87  &72.64 &73.43\\

 ~ & \checkmark  &\checkmark &52.62 &\textbf{63.99} &\textbf{68.21} &\textbf{70.19} &\textbf{71.92} &\textbf{72.79} &\textbf{74.15}\\
\hline
\end{tabular}
\end{center}
\end{table*}

\section{Experiments}
\label{exp}
\subsection{Experimental Settings}
{\bf Datasets.} 
The experiments use two datasets: PASCAL VOC dataset and MS-COCO dataset. For PASCAL VOC, it contains 20 categories, VOC07 trainval and VOC12 trainval are used in the training phase, and VOC07 test is used in the test phase. The number of images is 16551 for the training phase and active learning, and the test phase uses 4952 images. 
For MS-COCO, it contains 80 categories and includes lots of dense objects and small objects, which are difficult to object detection. For active learning, the training set uses MS-COCO 2014 which contains 117k images for training, and evaluates the results of MS-COCO 2017 which contains 5k images. The evaluation metric is mean average precision (mAP).

{\bf Target Models.}
For active learning, we mainly adopt the RetinaNet \cite{lin2017focal} based on ResNet-50 and SSD \cite{ssd} based on VGG-16 as the base detector, which is widely used in active learning research for object detection \cite{yoo2019learning,Yuan_2021_CVPR,choi2021active,wu2022entropy}. For RetinaNet on PASCAL VOC, we randomly selected 5\% of the training set as the initial labeled data. Then, in each cycle, we select 2.5\% of the training set from the rest labeled images until 20.0\% of the training set budget is consumed. The selection process reference \cite{Yuan_2021_CVPR}. For SSD on PASCAL VOC, we use 1000 labeled images as the initial labeled images. In each cycle, we also select 1000 images to add to the labeling image set, which refers to \cite{yoo2019learning}. The selection interval is about 6\% of the training set. For RetinaNet on COCO, the training configuration is the same as PASCAL VOC. For active learning, the selection interval is about 2\% of the training set. 

\begin{table}[t] 
\centering
\caption{Performance under the different number of selected instances for PASCAL VOC using RetinaNet.}
\begin{center}
\begin{tabular}{p{0.5cm}<{\centering}|p{0.6cm}<{\centering} p{0.6cm}<{\centering} p{0.6cm}<{\centering} p{0.6cm}<{\centering} p{0.6cm}<{\centering}p{0.6cm}<{\centering}p{0.6cm}<{\centering}}
\hline
\multirow{2}*{$Z$}  & \multicolumn{7}{c}{ mAP (\%) on Proportion (\%) of Labeled Images.}  \\
\cline{2-8}
~ & 5.0 &7.5& 10.0& 12.5 & 15.0 & 17.5 &20.0\\
\hline
1  &29.09 &51.94 &60.45 &62.66 &66.71  &67.13 &68.73 \\
10 &29.09 &49.46 &60.65 &63.88 &67.14 &68.48 &68.75\\
100&29.09 &48.92 &60.23 &64.47 &66.85 &68.82 &70.02\\
1k &29.09 &51.36 &61.82 &65.92  &67.86&69.41 &71.12\\
10k &29.09 &\textbf{53.43} &\textbf{63.98} &\textbf{66.96} &\textbf{69.73} &\textbf{70.37} &\textbf{71.41}\\
100k &29.09 &53.06 &63.17 &65.94 &67.26  &70.05 &71.09 \\
\hline
\end{tabular}
\end{center}
\label{tab2}
\end{table}

\begin{table}[t] 
\centering
\caption{Performance under the different number of members for PASCAL VOC using RetinaNet.}
\begin{center}
\begin{tabular}{p{0.5cm}<{\centering}|p{0.6cm}<{\centering} p{0.6cm}<{\centering} p{0.6cm}<{\centering} p{0.6cm}<{\centering} p{0.6cm}<{\centering}p{0.6cm}<{\centering}p{0.6cm}<{\centering}}
\hline
\multirow{2}*{$N$} & \multicolumn{7}{c}{ mAP (\%) on Proportion (\%) of Labeled Images.}  \\
\cline{2-8}
~ & 5.0 &7.5& 10.0& 12.5 & 15.0 & 17.5 &20.0 \\
\hline
1 &29.09 &49.54 &58.74 &61.91 &62.57 &65.27 &66.87 \\
2 &29.09 &52.78 &63.52 &66.71&68.35 &69.37  &71.21\\
3 &29.09 &53.43 &\textbf{63.98} &\textbf{66.96} &\textbf{69.73} &\textbf{70.37} &\textbf{71.41}\\
4 &29.09&50.41 &61.94 &66.68 &68.75 &69.29&70.29\\
5 &29.09 &\textbf{54.81} &61.21&63.64 &67.77 &69.82&71.11\\
6 &29.09 &48.93 &60.68 &65.51& 67.08&69.75&68.67\\
\hline
\end{tabular}
\end{center}
\label{tab3}
\end{table}

\begin{table}[t] 
\centering
\caption{Performance under the different $\lambda$ for PASCAL VOC using RetinaNet.}
\begin{center}
\begin{tabular}{p{0.5cm}<{\centering}|p{0.6cm}<{\centering} p{0.6cm}<{\centering} p{0.6cm}<{\centering} p{0.6cm}<{\centering} p{0.6cm}<{\centering}p{0.6cm}<{\centering}p{0.6cm}<{\centering}}
\hline
\multirow{2}*{$\lambda$} & \multicolumn{7}{c}{ mAP (\%) on Proportion (\%) of Labeled Images.}  \\
\cline{2-8}
~ & 5.0 &7.5& 10.0& 12.5 & 15.0 & 17.5 &20.0 \\
\hline
0.1  &29.09 &\textbf{54.43} &61.56 &65.35 &68.54 &\textbf{71.01} &\textbf{72.07}\\
1  &29.09 &53.43 &\textbf{63.98} &\textbf{66.96} &\textbf{69.73} &70.37&71.41\\
10 &29.09 &51.82 &62.78 &65.27 &68.26 &69.78 &71.23\\
\hline
\end{tabular}
\end{center}
\label{tab31}
\end{table}

\begin{table}[t] 
\centering
\caption{Performance under the different $\gamma$ for PASCAL VOC using RetinaNet.}
\begin{center}
\begin{tabular}{p{0.5cm}<{\centering}|p{0.6cm}<{\centering} p{0.6cm}<{\centering} p{0.6cm}<{\centering} p{0.6cm}<{\centering} p{0.6cm}<{\centering}p{0.6cm}<{\centering}p{0.6cm}<{\centering}}
\hline
\multirow{2}*{$\gamma$} & \multicolumn{7}{c}{ mAP (\%) on Proportion (\%) of Labeled Images.}  \\
\cline{2-8}
~ & 5.0 &7.5& 10.0& 12.5 & 15.0 & 17.5 &20.0 \\
\hline
0.1 &29.09 &\textbf{56.66} &61.12 &66.13 &68.22 &70.23 &71.33 \\
1  &29.09 &53.43 &63.98 &\textbf{66.96} &\textbf{69.73} &70.37 &\textbf{71.41}\\
5 &29.09 &55.82 &\textbf{64.73} &65.27 &68.35 &\textbf{70.83} &71.05\\
\hline
\end{tabular}
\end{center}
\label{tab32}
\end{table}

{\bf SOTA Methods.}
The proposed method is compared with Random sampling, Entropy sampling\cite{settles2008curious}, Core-set \cite{sener2017active}, LLAL \cite{yoo2019learning}, CDAL \cite{agarwal2020contextual}, GMM \cite{choi2021active}, MI-AOD \cite{Yuan_2021_CVPR}, and DivProto \cite{wu2022entropy}. For Entropy sampling, the sum of information entropy of all instances is the uncertainty of the image.
\begin{figure*}[htp]
\centering
\includegraphics[scale=1.2]{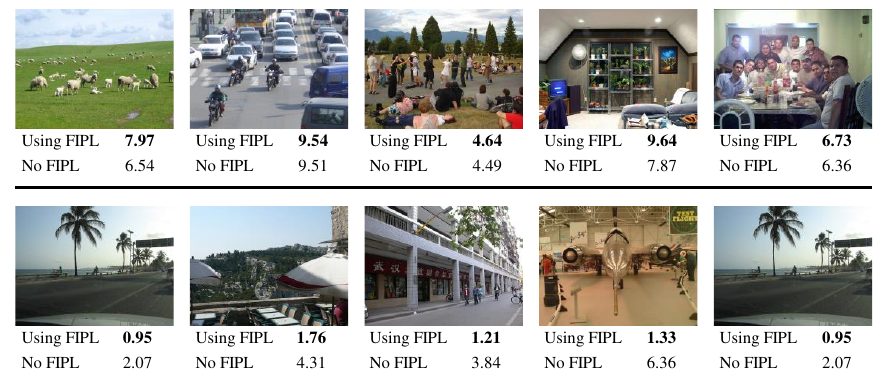} 
\caption{The uncertainty values of different images are displayed using Classification Committee. The unit is $10^{-6}$. The top: Ours. The bottom: Entropy sampling.} 
\label{fig3}
\end{figure*}

{\bf Implementation Details.}
All experiments are repeated 5 times, and the average result of 5 times is taken as the final experimental result. The $\lambda$ in Eq.(\ref{gs11}) is set 1, the $Z$ in Eq.(\ref{gs12}) is 10K, the $N$ in Eq.(\ref{gs5}) is 3, and the $\gamma$ is 1 in Eq.(\ref{gs10}). All experiments use an RTX 3080 GPU. Specifically, in Figure \ref{fig1}, the configuration of each experiment is as follows: For RetinaNet on PASCAL VOC, in each active cycle, the total number of epochs is 26, which contains 22 epochs for labeled images and 4 epochs for unlabeled images with the mini-batch size 2 and the learning rate 0.001. The momentum and the weight decay are set to 0.9 and 0.0001 respectively. For SSD on PASCAL VOC, in each active cycle, the labeled images is used to train the model 240 epochs and the unlabeled images is used to train the model 32 epochs. The momentum and the weight decay also are set to 0.9 and 0.0001. For RetinaNet on COCO, the training configuration is the same as PASCAL VOC.
\begin{figure*}[htp]
\centering

\includegraphics[scale=0.9]{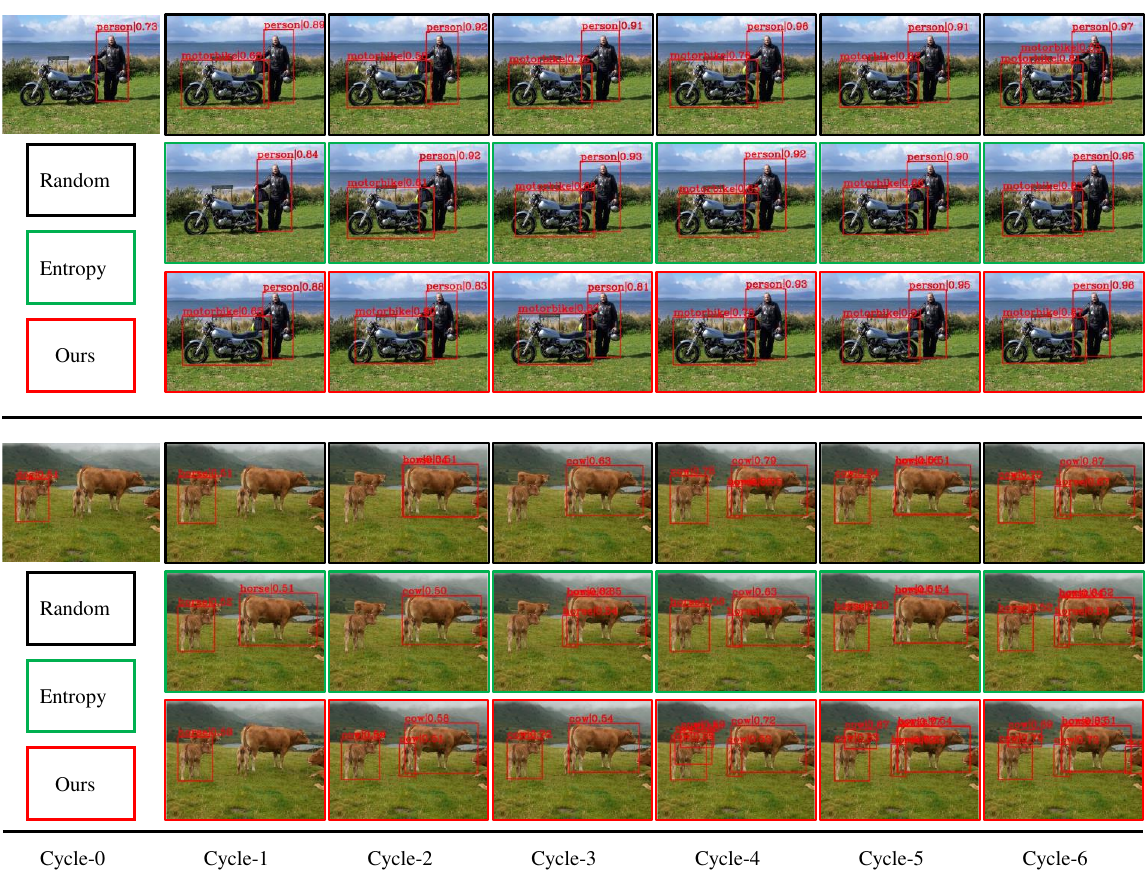} 

\caption{Compare the detection results with  Random and Entropy strategies on PASCAL VOC using RetinaNet.}
\label{fig2}

\end{figure*}
\subsection{Comparsion with SOTA Methods}
{\bf PASCAL VOC.}
Figure \ref{fig1} (1) gives the performance of our method compared with other methods for RetinaNet on PASCAL VOC. Obviously, our method vastly outperforms Random sampling, Entropy sampling, Core-set, CADL, LLAL, and DivProto in all cycles. Especially in the early stage of image selection, the increase in this period is larger than that in the later stage. The experimental result respectively outperforms state-of-the-art methods by 5.22\%, 5.47\%, and 2.86\% when selecting 5.0\%, 7.5\%, and 10.0\% images. 
Using up all the budget can reach 71.41\%, which significantly outperforms CDAL by 2.11\%. The experimental results of SSD are shown in Figure \ref{fig1} (2). The experimental effect of SSD also exceeds that of all comparison methods in all cycles. Compared with other methods, the degree of improvement in the first few cycles of image selection is obviously greater than that in the later cycles, and this phenomenon is similar to that when using RetinaNet. The experimental performance outperforms state-of-the-art methods by 1.17\% and 1.35\% when active learning is carried out in the second and third cycles respectively. 
Particularly, for MI-AOD, the method is a combination method of semi-supervised, active learning, and pseudo-labeling. Our method also Significantly outperforms MI-AOD in all selection cycle.


{\bf COCO.} 
In Figure \ref{fig1} (3), the performance of COCO is better than other methods. In the first cycle, it outperforms state-of-the-art methods by 1.14\%. Similarly to the previous experiments, the degree of improvement in the first few cycles of image selection is obviously greater than that in the later cycles. Analogously, our method also exceeds MI-AOD which combines semi-supervised and pseudo-labeling techniques. These comparative experiments show the general applicability of our method.
\begin{table}[t] 
\caption{Comparison of time cost on PASCAL VOC using RetinaNet.}
\begin{center}
\centering
\begin{tabular}{p{1.6cm}<{\centering}|p{0.457cm}<{\centering} p{0.457cm}<{\centering} p{0.457cm}<{\centering} p{0.457cm}<{\centering} p{0.457cm}<{\centering}p{0.457cm}<{\centering} p{0.457cm}<{\centering}}
\hline
\multirow{2}*{\small Methods}  &\multicolumn{7}{c}{\small Time (h) on Proportion (\%) of Labeled Images.} \\
\cline{2-8}
 ~ &5.0 &7.5 &10.0 &12.5 &15.0 & 17.5 & 20.0 \\
\hline
\footnotesize Random &2.13 &4.17 &9.62 &15.32 &22.73 &30.28 &40.92\\

\footnotesize {DivProto\cite{wu2022entropy}} &2.09 &9.26 &18.15 &31.87 &43.39 &57.23 &73.78\\
\footnotesize Ours& 2.61 &\textbf{6.96} &\textbf{12.52} &\textbf{19.31} &\textbf{27.55} &\textbf{36.73} &\textbf{47.14}\\
\hline
\end{tabular}
\end{center}
\label{tab5}
\end{table}

\subsection{Ablation Study}
{\bf MCDGL and FPIL.}
Table \ref{tab1} mainly shows that the ablation experiments of the most important module use two detection models. Apparently, the role of FIPL can further improve the accuracy of detection. Using two models for the same ablation experiments proves the effectiveness of MCDGL for discrepancy learning and FPIL for filtering interference instances and highlighting positive instances of uncertainty. 

{\bf Hyper-paramenters and Time Cost.}
Table \ref{tab2} shows the performance when $Z$ is set to different quantities in Eq.(\ref{gs12}) using RetinaNet for PASCAL VOC. We can observe that the overall performance is slightly better when using 10K selected instances. Table \ref{tab3} shows the performance when $N$ is set to a different number in Eq.(\ref{gs3}) using RetinaNet for PASCAL VOC. One classifier adopts entropy sampling. The results of using three members are more stable than others. Table \ref{tab31} shows the performance when $\lambda$ is set to a different number in Eq.(\ref{gs11}) using RetinaNet for PASCAL VOC. We can observe some differences in the performance, when $\lambda$ is set to 1, it looks slightly better overall. In terms of final performance, when $\lambda$ is set to 0.1, the accuracy can exceed 72\%. Similarly, table \ref{tab32} shows the performance when $\gamma$ is set to a different number in Eq.(\ref{gs10}) using RetinaNet for PASCAL VOC. There are some differences in experimental results, when $\gamma$ is set to 1, it looks slightly better overall until the end of selection. Table \ref{tab5} shows that our method takes less time, mainly because DivProto spends lots of time on information evaluation.

\begin{table}[t] 
\begin{center}
\caption{The number of true positive instances selected in each cycle on PASCAL VOC using RetinaNet.}
\centering
\begin{tabular}{p{1.7cm}<{\centering}|p{0.4cm}<{\centering} p{0.4cm}<{\centering} p{0.4cm}<{\centering} p{0.5cm}<{\centering} p{0.5cm}<{\centering} p{0.5cm}<{\centering} p{0.5cm}<{\centering}}
\hline
\multirow{2}*{\small Methods}  &\multicolumn{7}{c}{\small Number on Proportion (\%) of Labeled Images.} \\
\cline{2-8}
 ~ &5.0 &7.5 &10.0 &12.5 &15.0 & 17.5 & 20.0\\
\hline
\footnotesize Random &2405 &3586 &4829 &6038 &7105 &8368 & 9553\\
\footnotesize {Entropy\cite{settles2008curious}} &2405 &3537 &4784 &6076 &7253& 8451& 9637\\
\footnotesize Core-set \cite{sener2017active} &2405 &4234 &5471 & 6808 &8056 &9300 & 10607 \\
\footnotesize No FPIL& 2405 &6337 & 9319& 10762 &12709 &14848 &  16374\\
\footnotesize Ours& \textbf{2405} &\textbf{6514} &\textbf{9667} &\textbf{12213} &\textbf{14524} &\textbf{16730} &\textbf{18365}\\
\hline
\end{tabular}
\end{center}

\label{tab4}
\end{table}

\subsection{Additional Analysis}
{\bf Statistical Analysis.}
Table \ref{tab4} can be seen that the proposed method significantly selects more true positive instances in all learning cycles. Fewer positive instances are selected without FPIL than with FPIL. These selected instances are both positive instances and instances with high uncertainty. The more such instances are included in an image, the greater the information content of the image will be and the greater the improvement of the model will be, which also proves the effectiveness of our method.

{\bf Visual Display.} 
Figure \ref{fig2} shows the comparison of the object detection results of the random sampling method and entropy sampling method with the proposed method in this paper. The recognition accuracy of different objects is improved faster and the results are more stable. For densely stacked objects, this method can also be faster and more accurately separated. Obviously, the proposed method gets good results faster. 
In Figure \ref{fig3}, these discrepancy scores are obtained by the Classification Committee in the early cycle, the top half shows images with more objects by our method, and the bottom half shows images with fewer objects and more complicated background information by Entropy sampling. The results show that images with more objects have a larger uncertainty score, and images with complex backgrounds also have high uncertainty scores. 
Thus, sample selection is likely to pick up a lot of background information which prevents rapid model improvement. Using FPIL can significantly reduce the uncertainty of complex background images for selecting more valuable images. 
These phenomena directly prove the effectiveness of the proposed method. The experiment proves that our method can filter out the interfering instances, which selects more representative instances in an image for calculating uncertainty.

\section{Conclusion}
\label{con}
In the paper, we propose an active deep model for object detection, which contains a main detector and a classification committee. The main detector is the original detector, and the classification committee trained by Maximum Classifiers Discrepancy Group Loss (MCDGL) is as basic for uncertainty evaluation on object detection. To suppress interference instances, the Focus on Positive Instances Loss (FPIL) can modify MCDGL, which makes discrepancy learning more focused on positive instances. The sum of the selected uncertain instances of the images is the amount of information on the image. Experiments on different datasets and different object detection models have validated the superiority of our method compared with state-of-the-art methods. The proposed method can provide a new direction for active deep object detection. This paper's approach relies on uncertainty learning for active deep object detection. Future work will explore novel ideas for active learning, incorporating sample diversity measures into the learning process.

\end{document}